# Automatic Knowledge Graph Construction for Judicial Cases

Jie Zhou[1], Xin Chen[1], Hang Zhang[2], Zhe Li[1]
[1]School of Computer Engineering, Jiangsu Ocean University
[2] School of Law and Public Administration, Jiangsu Ocean University

**Abstract**

In this paper, we explore the application of cognitive intelligence in legal knowledge, focusing on the development of judicial artificial intelligence. Utilizing natural language processing (NLP) as the core technology, we propose a method for the automatic construction of case knowledge graphs for judicial cases. Our approach centers on two fundamental NLP tasks: entity recognition and relationship extraction. We compare two pre-trained models for entity recognition to establish their efficacy. Additionally, we introduce a multi-task semantic relationship extraction model that incorporates translational embedding, leading to a nuanced contextualized case knowledge representation. Specifically, in a case study involving a "Motor Vehicle Traffic Accident Liability Dispute," our approach significantly outperforms the baseline model. The entity recognition F1 score improved by 0.36, while the relationship extraction F1 score increased by 2.37. Building on these results, we detail the automatic construction process of case knowledge graphs for judicial cases, enabling the assembly of knowledge graphs for hundreds of thousands of judgments. This framework provides robust semantic support for applications of judicial AI, including the precise categorization and recommendation of related cases.

**0 Introduction**

In the judicial reform driven by artificial intelligence, facing the huge amount of adjudication documents and letting the machine recognize the cases through certain cutting-edge technology is the premise and weak point of the current judicial application of artificial intelligence. The realization of automatic machine learning and cognitive cases will have an important impact on a series of judicial applications, such as similar case retrieval, accurate delivery of class cases, and automatic generation of adjudication documents.

Currently, deep learning technology represented by connectionism and knowledge graph technology represented by symbolism are being widely and deeply studied, which will bring far-reaching impacts and changes to various industries. To this end, we use deep learning as the driving technology and knowledge graph as the knowledge carrier to realize the automatic construction of case knowledge graph for judicial cases, so as to realize the machine's cognition of cases.

The concept of knowledge graph was formally put forward by Google in 2012, and Google built the next generation intelligent search engine based on this technology. At present, the representative large-scale knowledge bases include: Freebase [1], Wikidata [2], dbpedia [3], Yago [4], Zhishi.me [5], CN-dbpedia [6] and so on. The above-mentioned knowledge base data are basically from the open community or open domain data, belonging to the general knowledge graph, which is of little significance to the actual vertical field application. With the upsurge of knowledge mapping research, the research of domain knowledge mapping has been paid more and more attention. For example, two large-scale open academic knowledge graphs OAG [7] and ACEKG [8] will be beneficial to the research and development of academic data mining. In addition, the construction and application of knowledge graph can also be seen in medical and financial fields.

At present, for vertical knowledge graphs, the data source is mainly structured text data, and the construction of knowledge graphs for unstructured text is not widely studied. For the unstructured text in vertical domains, it is not feasible to adopt the open information extraction method. To this end, we design a supervised entity recognition and relationship extraction pipeline model. For the entity recognition task, the current effective means is still based on deep learning methods, and the mainstream methods can be categorized into recurrent neural network RNN-based methods [9-10] (e.g., LSTM CRF), convolutional neural network CNN-based methods [11] (e.g., IDCNN), and hybrid models. IDCNN) and hybrid modeling approaches [12-13] (e.g. LSTM CNN CRF). For the task of relational extraction, effective and mainstream methods can still be categorized into RNN-based methods [14-15], CNN-based methods [16-18], and their hybrid models [19]. There are also rule-based methods [20-21] for (temporal) knowledge graph reasoning.

The above tasks generally utilize Word2Vec [22] with the emergence of pre-training modeling, the above problem has been solved. Some representative models include Based on two-layer bi-directional LSTM [23] The model of ELMo [24] based on the unidirectional Transformer [25] The model of GPT [26] and BERT [27], a model based on a bidirectional transformer and incorporating the next sentence prediction task. Unsupervised pre-training based on large-scale text can fully learn the semantic information embedded in it, which usually can directly improve the existing NLP tasks. For the entity recognition task, Google's BERT Softmax [29] has been used. The model [27] outperforms previous results; for the relational extraction task, a multi-task model using the pre-trained model GPT combined with a language model TRE [28] The best effect is achieved.

Judicial documents record the process and results of the people's court, compared with the Internet encyclopedia, news and information text, the characteristics of the judicial documents mainly include. The legitimacy of the document production, the text must be produced according to law, which is the basic premise; the form of the procedure, manifested in the structure of fixed and idiomatic culture; language accuracy, manifested in the semantic expression of a single, accurate and precise. Compared with the ruling, due to the large number of judgments, the facts of the case and the reasoning of the decision are more detailed, which is more valuable for the technical study of justice.

Judicial judgments mainly include structured case basic information and unstructured text. Structured case basic information reflects the subject of the case, which is also the basis of the facts of the case. Unstructured text type mainly includes the statement of the parties to the case, the court found the facts, the court reasoning and decision results of the three types of paragraphs, the statement of the parties to the case describes a certain objective facts, but due to a certain degree of subjectivity, there may be contradictory statements of the facts; the court reasoning and decision results are focused on legal norms based on the main text of the decision of the argument; the court found the facts, based on The court's reasoning and decision result focuses on the main text of the decision based on legal norms; the court's determination of facts, based on the situation of proof and cross-examination in the case, describes the facts that affect the decision result of the case. Therefore, based on the basic information of the case, it is very necessary and reasonable to construct the knowledge graph of the case around the text of the court's determination of the facts.

This paper takes the judicial judgment under the case of "motor vehicle traffic accident liability dispute" as the research object, and the research goal is to automatically construct a case knowledge graph for each document. The main contributions of this paper are as follows.

(1) We compare two BERT-based entity recognition models, and experimentally show that the use of CRF in the decoding output layer can further improve the entity recognition effect by 0.36. We also propose a fusion model that can be used to recognize entities in the case of traffic accident liability disputes.

(2) We propose a joint multi-task semantic relationship extraction model BERT Multitask that incorporates translational embedding, and compared with the baseline model, the relationship extraction result $F_1$ value is improved up to 2.37.

(3) We design an automatic process of constructing a case knowledge graph that integrates structured and unstructured texts, and the results verify the feasibility and effectiveness of the process, and construct a largescale case knowledge graph of judicial cases, which provides semantic support for downstream tasks such as accurate pushing of class cases.

**1 Entity Recognition Model**

Entities are presented in the form of nodes in the knowledge triad, which is the main body and foundation of the knowledge graph. The baseline model of entity recognition can be divided into three major network layers, namely, input embedding layer, feature extraction layer and decoding output layer.

There is a problem with the above benchmark model. The upper decoding prediction output layer adopts Softmax, and the predicted output labels of the sequence are independent of each other. In fact, there is a certain dependency between the output label sequence of entities. For example, the output label IMV can only follow the label BMV, but not the label BNP, where MV represents the motor vehicle entity and NP represents the natural person entity. Therefore, we use conditional random field (CRF [31]) as the decoding output layer to solve this problem.

The modified model BERTCRF, which differs from the baseline model in the decoding output layer. Combining the implied state output obtained from the above feature extraction $H$, define the composite score function as shown in Eq (2).

$$f(H,Y) = \sum_{i=1}^{n} A_{y_i y_{i+1}} + \sum_{i=1}^{n} P_{i,y_i}$$

Where $A$ is the transfer score matrix between the output labels, where, the $A_{ij}$ Corresponding labels $i$ to the label $j$ The score of the For the input implicit state $H$ Applying Softmax to all possible output label sequences, we obtain the predicted label sequence $Y$ The probability of the p-value, as shown in Eq. (3).

$$p(Y \mid H) = e^{f(S,Y)} / \sum_{\tilde{Y} \in Y_X} e^{f(S,\tilde{Y})}$$

We need to maximize the composite score, which is generally taken as the logarithm of the probability of the predicted output label sequence, as shown in Eq. (4).

$$\log(p(Y \mid H)) = f(H,Y) - \log\left(\sum_{\tilde{Y} \in Y_X} e^{f(S,\tilde{Y})}\right)$$

where, in the case of $Y_X$ represents the space of all possible output label sequences corresponding to the input implicit states.

According to the above formula, the output label sequence corresponding to the maximum score is the optimal predicted label sequence, as shown in Eq. (5).

$$Y^* = \underset{\tilde{Y} \in Y_X}{argmax} f(H, \tilde{Y})$$

Generally, only the transfer relationship between any two tags is considered. The above optimal solution can be obtained by dynamic programming, and we use Viterbi algorithm (Viterbi algorithm [32]) to decode it.

## 2 Relational Extraction Model

Inspired by the GPT model, introducing the language model as an auxiliary objective can improve the generalization performance of the model and speed up the harvest. We introduce the translation embedding (translating embedding [33], TransE) task of knowledge triples as an auxiliary optimization objective.

Fig. 1 illustrates a semantic relation extraction model BERT-Multitask that combines the tasks of relation classification and translational embedding. For each token in the input, initial embedding and feature extraction are performed to obtain a higher-level, more semantically rich implicit state H

$$H = (H_{[CLS]}, H_1, \cdots, H_n)$$

After obtaining the implicit state $H$ On the basis of [CLS], the features of sentence, entity 1 and entity 2 are further mapped and processed to get the features of sentence, entity 1 and entity 2 to prepare for the next step of feature fusion and multi-task joint learning.

For obtaining sentence features, take the implicit state $H_{[CLS]}$ of the first token (i.e., [CLS]), input it into the Pooler layer, and then we can get the feature representation of the sentence, as shown in Eq (6).

$$F_{sent} = tanh(H_{[CLS]} W_{sent})$$

where $W_{sent}$ is a weight parameter, which is mapped to the same dimension.

For the acquisition of entity features, we average the implied state outputs of all token sequences composing an entity as the feature representation of the entity, as shown in Eq. (7).

$$F_{ent} = \frac{1}{N} \sum_{k}^{N+k-1} H_i$$

where, the $N$ is the length of the sequence of the entity, the $k$ is the position index of the start of the entity sequence.

We concatenate sentence features, entity 1 features, and entity 2 features, input them into a feedforward neural network, and perform feature fusion to obtain the fused features $F_{fused}$, as shown in Eq (8).

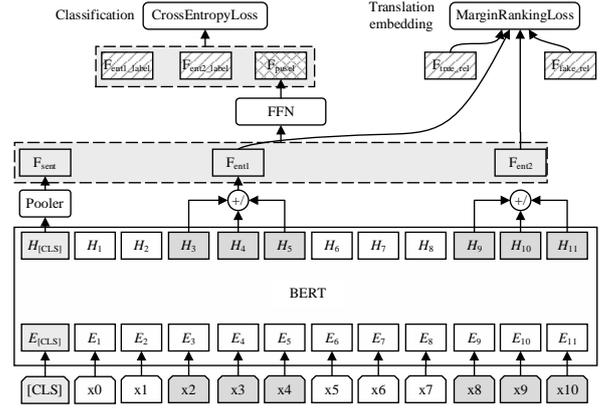

Fig. 1 A multitask federated semantic relation extraction model

$$F_{fused} = gelu\left((F_{sent} \oplus F_{ent\,1} \oplus F_{ent\,2})W_{fused} + b_{fused}\right)$$

where $W_{fused}$ is the weight parameter, the $b_{fused}$ is the bias parameter.

For the relationship classification task, considering that a specific relationship generally occurs only between two specific entity categories, we join the category features of entity 1, the category features of entity 2 and the hybrid features obtained above to obtain the final features $F_{final} = F_{ent\,1\,abel} \oplus F_{ent2\,aba\,l} \oplus F_{fused}$, input to the output layer Soft max, for output labeling $y$ is predicted as shown in Eq. (9).

$$P(y \mid F_{final}) = softmax(F_{final} W_{final})$$

In the relational classification task, the loss function Cross Entropy Loss is used to obtain the minimization optimization objective as shown in Eq (10).

$$L_1 = -\sum_{F_{final} \in F_X} log\, P(y \mid F_{final})$$

where, X is the input space, $F_X$ is the final feature space.

It is assumed that the features of the true relational labels are $F_{true\_el}$, the pseudo-relational labels are characterized by $F_{fake\_el}$, the loss function Margin Ranking Loss is used to obtain the minimization optimization objective for the translational embedding task, as shown in Eq (11).

$$L_2 = \sum_X \left[\gamma + d(F_{ent1} + F_{true\_el}, F_{ent2}) - d(F_{ent1} + F_{fake\_rel}, F_{ent2})\right]_+$$

where the distance d is measured using the $L_1$ norm as $d(F_{ent} + F_{rel}, F_{ent2}) = \| F_{ent\,1} + F_{rel} - F_{ent2} \mid 1, [x]_+ = max(x, 0), \gamma > 0$ is the interval hyperparameter to be optimized.

Considering the relational classification task and the knowledge triad translation embedding task together, the integrated loss function is obtained, as shown in Eq (12).

$$L = L_1 + \lambda * L_2$$

where $\lambda > 0$ is the weight coefficient for the loss of the translation embedding task.

## 3 Completion Method for Multi-Semantic Relational Embedding

The core of Knowledge Graph Completion lies in complex relationship reasoning, and choosing a knowledge representation model that can handle complex relationships is the basis for ensuring the complementation effect. The RotatE-based relational complementation model can deal with complex relations, but there is a problem of multi-semantic relations. In order to solve this problem, the following explorations are attempted.

(1) Does the RotatE representation model have the phenomenon of relational multisemantics? (2) How to subdivide the relational semantic representation? (3) How to ensure the uniqueness of vector computation? (4) Does the expanded semantic representation have the ability to handle complex relations? (5) How to verify the effect of knowledge graph complementation?

### 3.1 RotatE Model

The assumption space of RotatE model originates from the complex vector space of Complex, which is subject to Euler's constant equation $e^{i\pi} = \cos x + i\sin x$ inspired by the separation of entity and relation representations. Consider the relation as the head entity $h$ to the tail entity $t$ The process of rotation of i.e. $t = h°r$, the relation vector $r$ is denoted as $r = \begin{pmatrix} \cos\theta_r & -\sin\theta_r \\ \sin\theta_r & \cos\theta_r \end{pmatrix}, |r| = 1, \theta_r$ is the value of the angle between the head and tail entities, which is initialized by RotatE as a normal distribution vector, and the score function is defined as

$$d_r(h, t) = |h°r - t|$$

RotateE is able to infer symmetric, antisymmetric, inverse and combinatorial relations simultaneously. Specific relationship types are defined as follows.

**Definition 1** If there exists $\forall x, y$

$$r(x, y) \Rightarrow r(y, x)(r(x, y) \Rightarrow \neg r(y, x))$$

Then the relation r is defined as a symmetric (antisymmetric) relation.

**Definition 2** If there exists $\forall x, y$

$$r_2(x, y) \Rightarrow r_1(y, x)$$

Then define the relation $r_1$ is the relation $r_2$ the inverse relation.

**Definition 3** If there exists $\forall x, y, z$

$$r_2(x, y) \cup r_3(y, z) \Rightarrow r_1(x, z)$$

Then relation $r_1$ is defined as the combination of relations $r_2, r_3$.

Although RotatE can handle the existing complex relations, the semantic representation of a relation as a specific vector is not sufficient, and the use of the RotatE and other rotational models for knowledge representation are prone to missing part of the semantic representation.

After training RotatE on FB 15K-237 dataset, we obtain the entity set E and relationship set R. The corresponding relationship vectors are computed using Eq. The corresponding relationship vectors of entity set are calculated by using Eq (2), and the feature vectors are compressed into two-dimensional vectors (feature1, feature2) after PCA dimensionality reduction, and the four relationship angle vectors of position, award winner, nationality, and sports_team_roster are constructed.

It is reasonable to assume that RotatE treats a relationship as a specific vector. Then the entity and relationship vectors obtained after RotatE training should be unique, i.e., the vectors between the head and tail entity $r$ is unique. The angles between pairs of entities under the same relationship obtained after RotatE training should not differ much, which should be represented as a cluster on the graph, whose center is the relationship vector $r$ values.

Analyzing the four relations, it is found that the same relation presents the aggregation phenomenon of multiple clusters, i.e., the same relation has multiple vector representations, which is uniquely contradictory to the original assumption of vectors. The vector space constructed by Eq. (1) in RotatE can be regarded as an average of multiple clusters, but due to the large differences between semantic clusters, it leads to misclassification of most of the semantics in relationship prediction. Thus, it is clear that RotatE's representation of a relationship as a single vector does not adequately express the semantic information of the relationship.

### 3.2 MSRE method

In order to expand the original semantic representation of relations, a multi-semantic relation embedded knowledge graph completion method MSRE is proposed by replacing the relation vector representation with a relation semantic component smile, which can discover relations at multiple levels and dimensions through the semantic segmentation of relations.

MSRE (multi-semantic relation embedding) method is divided into three steps, as shown in Figure 2. (1) Calculate relationship semantic components. Firstly, the triad of knowledge graph is represented as entity and relation vectors, and then the relation vectors are transformed into a set of relation angle vectors. (2) Obtain relationship semantic component clusters. The Mean-Shift clustering algorithm is used to obtain the clusters of relationship angle vectors automatically. (3) Relationship Completion. Select the relationship vectors that are most appropriate to the entity pairs.

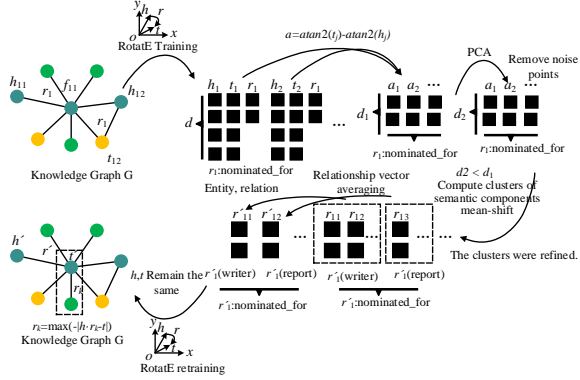

Fig. 2 MSRE method

### 3.2.1 Calculate the relationship semantic components

RotatE defines the relationship vector as the rotation angle between the head and tail entities, which is different from the embedding method of TransE, and adopts CTransR on the basis of TransE, through the $r = t - h$ The method of obtaining the set of relation vectors is not applicable. It is proposed to use RotatEtrained entity vectors to obtain each relation using the covariance difference, i.e., $r_i$ The set of angle vectors of angle $r_i = \{a_1, a_2, \cdots, a_c\}$, where $r_i \in \mathbf{R}$, $c$ denotes the number of components in the set of angles.

$$a = atan\ 2(t_j) - atan\ 2(h_j), a \in\ angle_{r_i}$$

where $t_j, h_j$ (dendte (jespectijyely the same relationship under $r_i$ of $j$ For head-tail entity pairs, the $a$ The range of values of $(-\pi, \pi)$.

### 3.2.2 Obtaining Relational Semantic Component Clusters

A relation semantic component family is a collection of semantic components under the same relation cluster angle $= \{sub_1, sub_2, \cdots, sub_k\}$, where $k$ denotes the number of semantic components. Since the semantic components of each relationship are discrete and numerous, a clustering algorithm is used to extract the main semantic components of the relationship semantic components. Comparing multiple clustering methods, we found that the traditional $K$ means algorithm requires manual setting of clusters, DBSCAN (densitybased spatial clustering of applications with noise) clustering method has poor clustering quality when the density of the sample set is not homogeneous and the difference in cluster spacing varies greatly, density-based Mean-Shift clustering algorithm shows better results in clustering in high-dimensional vector space, which is widely used in image segmentation, clustering, text classification and so on. Therefore, Mean-Shift algorithm is chosen to obtain semantic component clusters. In order to reduce the influence of invalid features, this paper carries out PCA (principal component analysis) dimensionality reduction before Mean-Shift, and averages the relationship semantic clusters obtained from clustering. Essentially the averaging process can be viewed as a relationship induced global information, helping the model to better collect roses.

Each relationship angle vector set angle $r_i$ is put into a high-dimensional vector space, and for any relational angle component in the space $b(b \in$ b(b ∈ angle $r_i$), the Mean-Shift vector is computed.

$$M_{g(b)} = \frac{1}{m} \sum_{x_i \in s_g} (x_i - b)$$

$s_g$ denotes a radius of $g$ The high-dimensional sphere region of the

$$s_{g(b)} = \{x \mid (x-b)^T(x-b) < g^2\}$$

T denotes the transpose, the $x_i$ is denoted as $s_{g(b)}$ A relational angle vector in the $m$ denotes the relational angle vector falling on $S_{g(b)}$ The number in the region, the $M_{g(b)}$ denotes the position of the next relational angle vector. After iteration, until $M_{g(b)}$ Collecting Mae. A new relational semantic component cluster angle is obtained $f = (r_{i1}', r_{i2}', \cdots, r_{ik}')$, $k$ denotes the relation $r_i$ The number of components in the Averaging the semantic relations of the same semantic cluster i.e:

$$v_n = \frac{1}{z} \sum_{j=1}^{z} d_j, n \in 1, 2, \cdots, k$$

$z$ denotes $r_{ij}$ Number of relation vectors in the same cluster. $d_j$ is the relational semantic component, cluster angle $e$ A semantic component in any one of the vectors. For all the classes in the cluster of relational semantic components after the computation, i.e. $r_i' = (v_1, v_2, \cdots, v_k)$, to obtain the set of relation vectors $R = (r_1', r_2', \cdots, r_k')$.

### 3.2.3 Relationship Completion

Representation learning-based complementation model first represents the entities and relationships in the knowledge graph in a low-dimensional vector space, whose forms include vector, matrix or tensor. Then, a ternary-based scoring function is defined on each knowledge entry, and the likelihood that a ternary or fact is valid is judged using the previously given knowledge graph representation, i.e., Threshold the set of entities E, the set of relations R with relations in the training knowledge graph to determine whether the current triad is valid or not.

In the MSRE method, the entity vectors in the knowledge graph are used as inputs, and each relation is represented by a cluster of semantic components, i.e., the set of relation vectors $R$, randomly construct negative examples that do not belong to the set of triples for training. After training, for a pair of entities to be judged, different relations and different semantic components of different relations, to obtain different scores. A higher score indicates that the entity pair is more likely to satisfy the relation.

$$d_r(\boldsymbol{h}, \boldsymbol{t}) = \max_{r_{ik}} - |\boldsymbol{h} \circ \boldsymbol{r_{ik}} - \boldsymbol{t}|$$

where $k$ denotes the number of classifications obtained after clustering.

In order to make the process of inducing clusters of relationship components not to reduce the accuracy of the model, this paper defines the selection strategy of relationships as the relationship vector that best matches the entity pairs, i.e., the relationship component with the highest score function is used as the relationship component in the ternary group.

Combined with the computational graph mechanism provided in Pytorch, the model can be expanded without decreasing the accuracy of the model. The generation strategy and loss function definition of the negative case ternary are consistent with the original RotatE model, and the adaptive relation vector representation of each relation component is obtained, and the model is trained by using the A dam optimizer. The model is trained using A dam optimizer, and when the model is relatively stable, a learning rate decay strategy, i.e., halving decay, is set to prevent the occurrence of overfitting and other problems. The specific algorithms are described as follows.

### 3.3 Theoretical Analysis

While expanding the semantic representation, the MSRE method also ensures the ability of the model to deal with complex relations, i.e., it does not destroy the original symmetric, antisymmetric, inverse, complex and other relations. The proofs are as follows.

(1) MSRE can deal with symmetric and antisymmetric relations, i.e., it does not destroy the original symmetric, inverse and complex relations.

$$r_k(x, y) \Rightarrow r_k(y, x)(r_k(x, y) \Rightarrow \neg r_k(y, x))$$

If $r_k(x, y), r_k(y, x)$ are proved, then $y = r_k \circ x \wedge x = r_k \circ y \Rightarrow r_k \circ r_k = 1$; If $r_k(x, y), \neg r_k(y, x)$, there is $y = r_k \circ x \wedge x \neq r_k \circ y \Rightarrow r_k \circ r_k \neq 1$.

(2) MSRE can handle inverse relations.

$$r_{2k}(x, y) \Rightarrow r_{1k}(y, x)$$

If $r_{2k}(x, y), r_{1k}(y, x)$ are proved, then $x = r_{1k} \circ y \wedge y = r_{2k} \circ x \Rightarrow r_{1k} = r_{2k}^{-1}$.

(3) MSRE can handle the combinatorial relation

$$r_{2k}(x, y) \wedge r_{3k}(y, z) \Rightarrow r_{1k}(x, z)$$

If $r_{2k}(x, y), r_{3k}(y, z), r_{1k}(x, z)$ are proved, then $z =$ $r_{1k}^\circ x \wedge y = r_{2k}^\circ x \wedge z = r_{3k}^\circ y \Rightarrow r_{1k} = r_{2k}^\circ r_{3k}^\circ$

## 4 Experiments
### 4.1 Data preparation
#### 4.1.1 Predefined entities and relationships

We take the first instance judgment of the civil case "Motor Vehicle Traffic Accident Liability Dispute" as the object of study, organize legal experts from universities and relevant judicial enterprises to participate in the research and discussion, and predefine the entities and relationships involved in the case in combination with the actual documents and existing legal norms.

In the end, we predefined 20 types of entities under the case. Considering that there are too many types of violations, and most of them appear less frequently in the documents, we chose 9 types of violations that are more common, and their numbers are corresponding to $12 \sim 20$.

There exist nine predefined relationship types, and a relationship type may correspond to multiple entity type pairs. For example, for the "Driving" relationship numbered 1, there are two conceptual knowledge triples (natural person subject driving a motorized vehicle) and (natural person subject driving a non-motorized vehicle), and a total of 30 conceptual knowledge triples (not counting the "other" relationship types). The total number of conceptual knowledge triples is 30 (not counting the "other" relationship category).

#### 4.1.2 Data Preprocessing and Labeling

Considering the differences in the level of economic and technological development among the eastern, central and western regions of China, there will be some differences in the level of judicial capacity and the standard of writing documents. We select "Jiangsu Province" and "Zhejiang Province" in the east, "Henan Province" and "Hubei Province" in the center, and "Sichuan Province" in the west. In the east, "Jiangsu" and "Zhejiang" were selected, in the center, "Henan" and "Hubei" were selected, and in the west, "Sichuan" and "Yunnan" were selected in order to address the effects of geographical variation. In each province, 100 judgments were randomly selected, and a total of 600 judgments were obtained as the original instrument data for the study.

The original judgments were pre-processed. First of all, we need to label the paragraph types of the documents, considering the structural norms and idiomaticity of the documents, we adopt a rule-based method to label the paragraph types of the documents. In this paper, the text paragraph types of the knowledge graph construction source include "party information" and other types of structured text and "the court found the facts" type of unstructured text, we randomly select 500 judgments to be pre-labeled with rules and submit them to manual review, and evaluate the effect of rule-based segmentation on these two types of texts. $F_1$ values are 99.85 and 90.34, respectively.

Based on the extracted structured text, we utilize a rule-based approach to extract the civil subjects involved to obtain the basic information of the case and the "plaintiff" and "defendant" in the facts of the case for completing the processing, and for the subsequent entity alignment.

We use an open-source tagging tool BRAT [34] to deploy and configure it, so as to realize online tagging of entities and relationships by many people, and treat the factual texts recognized by the court in clauses and import them into the tagging system.

After manual labeling and reviewing, and removing cases with too short length and poor quality, we finally obtained the labeled texts of 585 cases with facts found by the court. In order to avoid the influence of sentence-level division of data on the objective evaluation, we divided the data set at the case level, and the data set is divided, and the data volume statistics of entity task and relationship task are for the sentence level. When building the relational task dataset, for two entities that do not have a relationship and are within the definition of the relational entity type pairs, the "Other" relationship label is selected and the "Negative Sample" is taken as the "Negative Sample" with a 0.5 retention probability.

### 4.2 Entity Recognition

This experiment is conducted in a Tesla T4 16GB GPU environment, using the PyTorch framework for development, under the premise of basically no loss of precision, in order to reduce the memory overhead of the GPU, accelerate the training speed, we incorporate a mixed precision training technique called apex (1) into the programming, and the BERT model uses the implementation of PyTorch (2) (the task of relationship extraction is consistent with this).

Some important hyper-parameter settings involved in this experiment are mainly debugged based on previous work and practical experience, and no rigorous mesh search is performed. The parameters of the BERT-Softmax and BERTCRF models are not consistent except for the initial learning rate, which is $2E-5$, the latter being $1E-5$ The maximum length of the sentence is set to 400, and the sentences exceeding this length are divided into two segments by punctuation; an L2 regularization term with a coefficient of 0.01 is added to the weight parameter (the bias term is not processed), and the dropout [36] of the linear layer at the top layer is set to 0.1 to avoid over-fitting; The small batch size is set to 16; The threshold of gradient clipping is set to 2.0.

Compared with the benchmark model BERT-Softmax, the corrected model BERTCRF has a decrease of 0.27 in accuracy, an increase of 1.01 in recall, and an increase of 1.01 in the overall metrics on the test set. $F_1$ There is an improvement of 0.36. Taken together, the modified model outperforms the baseline model, showing that incorporating the transfer constraints between the output labels can further improve the entity recognition.

Based on the better BERTCRF model, the number of entities in each entity category in the test set and the corresponding F$F_1$ The values are shown in Fig. 5. Since the number of occurrences of the nine common illegal entity categories is less than 20, they are not involved in the statistics. The statistical results show thatin the 11 categories of entities with the number of entities greater than or equal to 20, F$F_1$ There are 6 categories with values above 95,7 categories with values above 90, and 8 categories with values above 85, and the overall performance is good. However, the performance of non-natural person subjects, non-motorized vehicles and property damage compensation program entities is poor. Analyzing the reasons, first of all, the number of these three types of entities is relatively small, the data is insufficient, resulting in insufficient model learning; the number of entities in the death category is very small, but because of its expression, such as "death", "death" and so on is relatively fixed, and thus can also get better results; the second reason lies in the diversity of entity expression, and the overall effect is good. The second reason lies in the diversity of entity expression, such as non-motorized vehicles and property damage compensation items expressed in a variety of forms, for example, property damage may involve a variety of items loss expression, resulting in more difficult to learn the model.

### 4.3 Relational Extraction

There are some important hyper-parameter settings in the relationship extraction task, and the meaning of some parameters with the same name is basically the same as that of the parameters introduced in the entity task. In particular, the embedding dimension of entity labels is set to 128, the embedding dimension of relation labels is set to 768, and the loss weight of the translation embedding task is set to $1E-5$, the interval parameter for the translation embedding task takes the default value of 1.0.

Referring to the official evaluation standard of SemEval-2010 Task 8 [37] multi-relationship classification task, we take the macro-averaged accuracy, recall rate and $F_1$ value to evaluate the effect. The only difference is that the academic research standard task evaluation does not consider the "other" relationship category. Considering that the model should be put into practical use and evaluated more objectively, we also include the "other" relationship category. Experiments show that the performance of "other" relationship extraction is often lower than the average.

Compared with the benchmark model BERT-Base, the improved model BERT-Multitask has higher accuracy, recall and F$F_1$ values have all gained an overall improvement, and the composite index $F_1$ The improvement is as high as 2.37, which indicates that the semantic relation extraction model incorporating translational embedding with multi-task association can significantly improve the effect of relation extraction, and verifies the effectiveness of incorporating semantic information constraints. After the training of BERTMultitask, a very valuable by-product can be obtained from the model, i.e., a vector embedding representation of entities and relations that combines contextual and ternary semantic relations.

Based on the better-performing BERT-Multitask model, the number and performance of each relationship category on the test set are shown in Fig. 6. Among the nine relationship categories, the $F_1$ There are 4 classes with values above 95, 6 classes with values above 90, and 7 classes with values above 85, and the overall performance is good. Negative sample of "other" relationship class $F_1$ The value is 91.05, which is lower than the overall performance of 91.64, while for the two types of relationships, "ride" and "accident", their extraction effect is a bit poorer, after analyzing, we found that the amount of data involving these two types of relationships is relatively small, and the distance between two entities of these two types of relationships tends to be farther away. After analyzing, we found that the amount of data involving these two types of relationships is relatively small, and the distance between the two entities of these two types of relationships is often far away.

### 4.4 Automatic construction of case knowledge graph

In the case of "motor vehicle traffic accident liability disputes", the text types constructed by the case knowledge graph include structured text and unstructured text. The paragraph types involved in the quasi-structured text include: document title, case number, court accepting and party information, while the unstructured text only contains the paragraph type of "court finding facts". The flow chart of automatic construction of case knowledge graph of judicial cases is shown in Figure 3.

For structured text, we use rule-based extraction to obtain the basic information of the case; for unstructured text, we use deep learning-based extraction to obtain the case relationship knowledge triad; the basic information of the case and the knowledge triad are fused to ultimately obtain the case knowledge graph of the case.

The detailed steps of constructing the case knowledge graph for judicial cases are as follows.

Step 1 Segment labeling. Given a judicial decision Doc, use the rule-based method to tag the segments, identify the structured text $Text_1$ and the unstructured facts of the case $Text_1$ as defined above, and then use the rule-based method to tag the case $Text_1$ and $Text_2$ as defined above;

Step 2 Class Structured Information Extraction. Rule-based extraction of $Text_1$ Extraction, to obtain the "title of the document", "case number", "receiving court" as the "case" entity's attributes information, extract the basic information of civil subject Info from the text of "party information" category, involving the name and information of its authorized agent, etc.;

Step 3 Data Preprocessing. $Text_2$ Pre-processing the text, which involves the original and the defendant's referential completion and clause processing, etc., to obtain the list of sentences $List_1$;

Step 4 Entity Recognition. Based on the trained entity recognition model BERTCRF, the entity recognition of $List_1$ Identify entities one by one, and get the list of entity data $List_2$, each piece of data contains. Each piece of data contains: a collection of sentences and entities (including categories)

Step 5 Relationship extraction. Combine the entities contained in each data item of $List_2$ within the range of predefined relational entity pairs, and form a relational data $list_3$ Each piece of data contains. Entity 1 and its category, Entity 2 and its category, the sentence in which it is located. The learned relational extraction model BERT-Multitask is utilized to analyze $List_2$ One by one, the relationship extraction is carried out, and finally the case fact triad Triplets is obtained.

Step 6 Knowledge fusion. Since the relation categories are standard predefined, the fusion mainly realizes the entity alignment of Info and Triplets. Some rule-based methods are mainly used. For example, rules are formulated by using fixed expressions between two entities, such as "hereinafter referred to as" and "abbreviated as"; According to the characteristics of the entities themselves, such as the entity categories and relationship constraints of "Sichuan AX A××××× Car" and "Sichuan A ××××××××××", the alignment can be made according to the license plate number. After entity alignment processing, case knowledge is obtained;

Step 7 Knowledge storage and visualization. Write Knowledge into Neo4j graph database for storage and visualization.

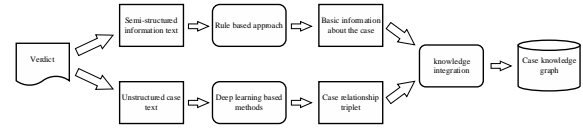

Figure 3: Flowchart of automatic construction of case knowledge graph.

For a newly input judicial judgment, the knowledge graph of the case automatically generated, which verifies the feasibility and effectiveness of the construction process. According to this process, we have selected more than 200,000 first instance judgments under the case of "motor vehicle traffic accident liability disputes" and automatically constructed the case knowledge graph, and obtained a large-scale case knowledge graph of judicial cases.

**5 Conclusion**

This paper explores the automated creation of case knowledge graphs for judicial cases, focusing on entity recognition and relationship extraction using advanced BERT-based models. We propose a method combining structured and unstructured texts to build large-scale knowledge graphs, proven effective through experimentation. Future efforts will refine these processes by incorporating supervised learning for paragraph type labeling and expanding data annotation to improve knowledge graph quality. We also aim to establish a robust evaluation system and explore the judicial applications of these knowledge graphs for precise case retrieval and analysis.